\documentclass[11pt,a4paper]{article}
\usepackage[hyperref]{emnlp2020}
\usepackage{times}
\usepackage{latexsym}

\usepackage{color}
\definecolor{mypink2}{RGB}{219, 48, 122}
\definecolor{jrcolor}{RGB}{100, 150, 225}
\definecolor{jrcomment}{RGB}{70, 200, 150}

\usepackage{amsmath, bm}
\usepackage{amsfonts}
\usepackage{multirow}
\usepackage{cleveref}
\usepackage{booktabs}

\crefformat{section}{\S#2#1#3}

\usepackage{microtype}

\usepackage{graphicx}
\usepackage[normalem]{ulem}

\aclfinalcopy 

\title{On the Branching Bias of Syntax \\ Extracted from Pre-trained Language Models}

\author{Huayang Li~~~Lemao Liu~~~Guoping Huang~~~Shuming Shi \\
        Tencent AI Lab\\
        \tt \{alanili,redmondliu,donkeyhuang,shumingshi\}@tencent.com 
}

\date{}

\begin{document}
\maketitle
\begin{abstract}
Many efforts have been devoted to extracting constituency trees from pre-trained language models, often proceeding in two stages: feature definition and parsing. However, this kind of methods may suffer from the branching bias issue, which will inflate the performances on languages with the same branch it biases to. In this work, we propose quantitatively measuring the branching bias by comparing the performance gap on a language and its reversed language, which is agnostic to both language models and extracting methods. Furthermore, we analyze the impacts of three factors on the branching bias, namely parsing algorithms, feature definitions, and language models. Experiments show that several existing works exhibit branching biases, and some implementations of these three factors can introduce the branching bias.
\end{abstract}

\section{Introduction}

Neural language models such as \textsc{LSTM}~\cite{merityRegOpt,peters2018deep}, \textsc{GPT2}~\cite{radford2019language}, and \textsc{BERT}~\cite{devlin2019bert,liu2019roberta} have achieved state-of-the-art performance in various downstream NLP tasks. Many recent works try to interpret their success by revealing the linguistic properties captured by these language models \cite{hewitt2019structural, clark2019does, jawahar2019does, tenney2019you}. One interesting line of these works tries to extract discrete constituency trees from pre-trained language models \cite{marevcek2018extracting,marevcek2019balustrades,kim2020pre, wu2020perturbed}. The core of these works is to extract syntax in two stages. Firstly, it defines the feature scores based on a language model, namely, the {\it feature definition stage}. Secondly, it leverages the feature scores to build a constituency tree, namely, the {\it parsing stage}.

However, the degree to which the extracted constituency trees match gold constituency annotations may imprecisely reflect the model's competence of capturing syntax, since their final performance may benefit from the branching bias.
For example, as pointed out by~\citet{dyer2019critical}, the syntax extracted from the ordered neuron based language model~\cite{shen2018ordered} is biased to right-branching languages \footnote{Right-branching language is considered to be head-initial, which means the head (e.g., the verb is the head in a verb phrase) always precedes its complements. In contrast, left-branching language is head-final, where the head follows its complements \cite{kiparsky1996shift,
kroch2001syntactic}.}  (e.g., English). Nevertheless, the approach to measuring the bias in~\citet{dyer2019critical} is highly dependent on the architecture of ordered neuron and its parsing algorithm. Therefore, it is far from trivial to be applied to general pre-trained language models.

\begin{figure}
  \begin{center}
      \includegraphics[width=6. cm]{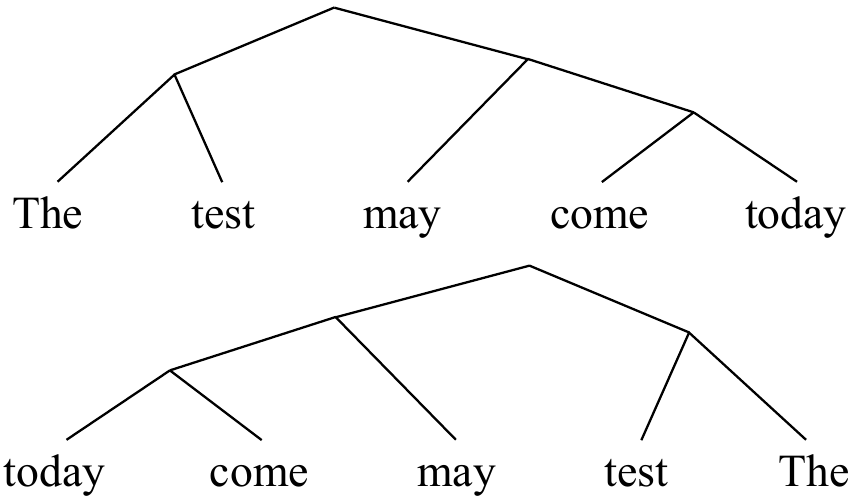}
      \caption{Constituency trees of a right-branching language and its reversed (left-branching) language. \label{fig:trees} The tree at the bottom is obtained by reversing the tree at the top.}
  \end{center}
\end{figure}

This paper proposes a new approach to reveal the branching bias of syntax from pre-trained language models, which is agnostic to model architectures and parsing algorithms. 
The key idea of our approach is based on the following observation: We can construct a left-branching language by reversing a right-branching language and vice versa. An illustration is given in Figure  ~\ref{fig:trees}. If a syntax extracting method has no branching bias, the parsing performances on the original language and the reversed language should have little or no difference.
Therefore, the performance gap can be used as an indicator of branching bias. 
Using our approach, we find that some recent works on pre-trained language models suffer from the branching bias~\cite{kim2020pre,wu2020perturbed, marevcek2018extracting}.
We further investigate on an in-depth question: Does the bias come from language models? Or the extraction methods (feature definition and parsing algorithm)?
We propose a simple approach to quantitatively analyze the bias in them, which tries to control the impacts of other factors while studying a specific part in the pipeline.

\section{Methodology\label{sec:methodology}}

\subsection{Measuring branching bias}

Intuitively, branching bias means that the induced syntax tends a specific branching structure, such as the right-branching in~\citet{shen2018ordered}, as pointed out by \citet{dyer2019critical}. For example, a right-branching bias will exaggerate the method's performance on a right-branching language and undermine its performance on a left-branching language. 
Therefore, a natural way to quantify the branching bias in syntax is to compare the performance gap between two natural languages with different branches (e.g., English and Japanese). 

Unfortunately, due to the intrinsic differences between the two natural languages, it may be unfair to compare their performances directly.  Therefore, for a language $L$, we build a synthetic language $L^\prime$ by reversing the word order in the way of right-to-left, rather than the left-to-right order in language $L$. If language $L$ is right-branching, then language $L^\prime$ will be left-branching, as shown in Figure \ref{fig:trees}.  Based on this observation, we use a natural language $L$ and a synthetic language $L^\prime$  to measure the performance gap. 

More concretely, the performance gap between language $L$ and $L^\prime$, namely the {\it branching gap}, is defined as follows:
\begin{equation}
  B = m(\boldsymbol{t}, \boldsymbol{g}) - m(\boldsymbol{t}^\prime, \boldsymbol{g}^\prime),\label{eq:branching_gap}
\end{equation}
where $m$ is a metric function to measure the quality of the parsing tree (e.g., f1-score); $\boldsymbol{t}$ is a tree extracted by a syntax extracting method on language $L$, and $\boldsymbol{g}$ is its golden truth; $\boldsymbol{t}^\prime$ and $\boldsymbol{g}^\prime$ are defined similarly but on the reversed language $L^\prime$. 
To make the comparison in Eq.\eqref{eq:branching_gap} fairer, we guarantee that training and testing datasets for both languages are the same except for the word order.

If a syntax extracting method is unbiased, the branching gap would be nearly 0.\footnote{Though we defined the branching gap on the sentence level, it can be easily applied to the corpus level.} The sign of indicates the direction of the branching bias. It is worth noting that the proposed approach to measure branching bias is independent of the model architecture and the syntax extracting method, unlike the approach used in~\citet{dyer2019critical}. Therefore, our approach can be naturally applied to any pre-trained language models and syntax extracting methods. Besides, \citet{dyer2019critical} mainly focus on the branching bias in a specific parsing algorithm~\cite{shen2018ordered}. In our work, we further analyze the branching bias in feature definitions and language models, besides parsing algorithms.

\subsection{Factors affecting branching bias\label{sec:bias_in_parser}}
Since constituency trees are extracted from a pre-trained language model using a syntax extracting method, 
the branching bias may owe to both the syntax extracting method and the language model. 
More precisely, the branching bias may be affected by  parsing algorithms, definitions of feature scores, and language models. 
In the rest of this section, we will investigate the branching bias in each of the three factors one-by-one. 



\begin{table*}[t]
  \begin{center}
  \begin{tabular}{c|c || c | c ||c|c|c}
  \# & Syntax Extracting Method & Model & Parsing Alg.  & $L$ & $L^\prime$ & Gap \\\hline
  1 &  \multirow{2}{*}{\cite{marevcek2018extracting}}  & \textsc{BERT} & \textsc{AttnSpan}  &   $27.81$    &   $29.60$   & $-1.79$  \\
  2 &   & \textsc{GPT2} & \textsc{AttnSpan}  &   $27.90$    &   $23.49$   & $+4.41$  \\\hline
  3 & \multirow{2}{*}{\cite{kim2020pre}}  & \textsc{BERT} & \textsc{Dist}  &  $24.93$    &   $25.23$   & $-0.30$  \\
  4 &  & \textsc{GPT2} & \textsc{Dist}  &  $31.09$    &   $36.29$   & $-5.20$  \\\hline
  5 & \multirow{2}{*}{\cite{wu2020perturbed}}  & \textsc{BERT} & \textsc{Mart}  &   $35.82$    &   $19.52$   & $+16.30$  \\
  6 &   & \textsc{GPT2} & \textsc{Mart}  &   $32.39$    &   $21.99$   & $+10.40$  \\

  \end{tabular}
  \end{center}
  \caption{\label{tab:pipeline} The branching gaps of some syntax extracting methods. The results are corpus-level F1 scores on English. $L$ is the original language and $L^\prime$ is its reversed version.}
  \end{table*}

\paragraph{Bias in parsing algorithm}
Since the parsing algorithm is on the top of the language model and feature definition,
To analyze the bias in a parsing algorithm alone, we need to exclude the influences of these two factors.
To this end, we propose to assign a sequence of random scores as the feature scores and then run the parsing algorithm using these random scores to obtain the constituency tree. The random feature scores are generated according to a uniform distribution\footnote{For feature scores that need to be normalized, and we will assign the random value before the normalizing operation.}.
Since the feature scores are independent of both the language model and the feature definition, the branching bias can be introduced solely by the parsing algorithm if a noticeable branching gap is observed.

\paragraph{Bias in feature definition}
Feature definition is the type of information (e.g., hidden vectors or attention matrix) from a language model, converted into feature scores, and then fed into a parsing algorithm.
Some feature definitions may also intrinsically contain branching bias. To reveal the bias solely dependent on a specific feature definition, instead of using the original weights (e.g., hidden representations and attention weights) outputted by a pre-trained language model, we propose randomly initializing them and using them to compute feature scores. Then we run an unbiased parsing algorithm on the feature scores generated in this way. Therefore, if there is a noticeable branching gap, the branching bias will be attributed to the feature definition. The pipeline to extract syntax is independent of the language model, and the fixed parsing algorithm is unbiased.


\paragraph{Bias in language model}
The pre-trained language model is the input of a syntax extracting method. We further analyze the branching bias in a language model. To analyze the branching bias in it, we firstly choose an unbiased syntax extracting pipeline (i.e., both the feature definition and parsing algorithm are fair) and then calculate the branching gap using the well-trained language models on languages $L$ and $L^\prime$. 
Since there is no branching bias within our selected extracting method, the branching bias can be attributed to the input itself, if a branching gap is observed. 


\section{Experiments}


\begin{table*}[t]
\centering
 \resizebox{2\columnwidth}{!}{
\begin{tabular}{{l|l|lllllllllll}}

\multirow{2}{*}{\#} & \multirow{2}{*}{Parsing Alg.}& \multicolumn{3}{c}{\centering {EN}} & & \multicolumn{3}{c}{ZH}  & & \multicolumn{3}{c}{DE} \\
 \cline{3-5}  \cline{7-9} \cline{11-13}
 &  & \multicolumn{1}{c}{$L$} & \multicolumn{1}{c}{$L^\prime$} & \multicolumn{1}{c}{Gap} & & \multicolumn{1}{c}{$L$} & \multicolumn{1}{c}{$L^\prime$} & \multicolumn{1}{c}{Gap} & & \multicolumn{1}{c}{$L$} & \multicolumn{1}{c}{$L^\prime$} & \multicolumn{1}{c}{Gap} \\ \hline

1 & \textsc{AttnSpan} &  $21.37$    &   $21.45$   & $-0.08$ & & $17.15$ & $16.98$ & $+0.17$ & & $17.79$ & $17.78$ & $+0.01$ \\
2 & \textsc{Dist} & $18.30$    &   $18.27$  & $+0.03$ & &  $15.28$ & $15.76$ & $-0.48$ & & $17.01$ & $16.94$ & $-0.07$ \\
3 & \textsc{Mart} & $26.11$   &  $15.41$ & $+10.70$ & & $19.90$ & $9.51$ & $+10.39$ & & $8.95$ & $17.07$ & $-8.12$ \\\hline
4 & \textsc{Random} & $18.31$  &   $18.37$  & $-0.06$  & & $15.33$ & $15.03$ & $+0.30$ & & $16.99$ & $16.98$ & $+0.01$ \\ 
5 & \textsc{Right-B} & $35.82$ & $10.40$ & $+25.42$  & & $19.77$ & $8.11$ & $+11.66$ & & $7.99$  & $16.54$ & $-8.55$ \\
\end{tabular}}
\caption{\label{tab:parser} The branching gaps of parsing algorithms with random feature scores. Feature scores are generated according to a uniform distribution $[-1, 1]$. The results are averaged corpus-level F1 scores with 10 random seeds. $L$ is the original language and $L^\prime$ is its reversed version.}
\end{table*}

\begin{table*}[t]
\centering
 \resizebox{2\columnwidth}{!}{
\begin{tabular}{{l|l|lllllllllll}}

\multirow{2}{*}{\#} &\multirow{2}{*}{Feature Def.}& \multicolumn{3}{c}{\centering {EN}} & & \multicolumn{3}{c}{ZH}  & & \multicolumn{3}{c}{DE} \\
 \cline{3-5}  \cline{7-9} \cline{11-13}
&  & \multicolumn{1}{c}{$L$} & \multicolumn{1}{c}{$L^\prime$} & \multicolumn{1}{c}{Gap} & & \multicolumn{1}{c}{$L$} & \multicolumn{1}{c}{$L^\prime$} & \multicolumn{1}{c}{Gap} & & \multicolumn{1}{c}{$L$} & \multicolumn{1}{c}{$L^\prime$} & \multicolumn{1}{c}{Gap} \\ \hline

1 & \textsc{Hidden} & $18.39$    &   $18.29$   & $-0.10$ & & $15.32 $&$ 15.30 $&$ +0.02 $ & & $ 16.88 $&$ 17.10 $&$ +0.28 $ \\
2 & \textsc{Prefix-Attn}  & $20.44$    &   $13.17$  & $+7.27$ & &  $ 16.78 $&$ 12.66 $&$ +4.12 $ & & $ 14.93 $&$ 18.83 $&$ -3.90 $ \\
3 & \textsc{Full-Attn}  & $18.33$   &  $18.38$ & $-0.05$ & & $ 15.12 $&$ 15.04 $&$ +0.08 $ & & $ 16.84 $&$ 16.79 $&$ +0.05 $ \\ 
\end{tabular}}
\caption{\label{tab:definition} The branching gaps while applying \textsc{Dist} to different randomized feature definitions. The uniform distribution $[-1, 1]$ is used to randomize the weights of feature definitions. The results are averaged corpus-level F1 scores with 10 random seeds.}
\end{table*}

\subsection{Settings}
\paragraph{Data} We choose English as the main language in our experiments. The English data used for training language models is the concatenation of 1M lines of Wikipedia data \cite{devlin2019bert} and the Penn TreeBank (PTB) \cite{Marcus:1993:BLA:972470.972475}  training data. We use PTB-22 and PTB-23 for validation and test, respectively. Besides, to rule out the impact of other linguistic properties, we also conduct part of our experiments on German and Chinese. We use the German Treebank from the SPMRL \cite{seddah-etal-2014-introducing} and Penn Chinese TreeBank (CTB) \cite{xue2005penn} with their provided test sets to evaluate previous methods on those two languages, respectively.

\paragraph{Language Models} In our experiments, we train three different language models (i.e., \textsc{BERT}, \textsc{GPT2}, \textsc{LSTM}) for English and its reversed language \footnote{We train a language model on the reversed language by reversing the entire training corpus}. The \textsc{BERT} and \textsc{GPT2} models are trained using Huggingface's Transformers \cite{Wolf2019HuggingFacesTS} and we use the default parameters of their base settings~\cite{devlin2019bert, radford2019language, Wolf2019HuggingFacesTS}. The \textsc{LSTM} model is trained using awd-lstm-lm\footnote{\url{https://github.com/salesforce/awd-lstm-lm}}, and we use the parameters similar to \citet{merityRegOpt}. Models used for extracting syntax are selected according to the PPL on validation set. The tokenizers for \textsc{BERT} and \textsc{GPT2} are trained using the toolkit huggingface/tokenizers\footnote{\url{https://github.com/huggingface/tokenizers}}, and their vocabulary sizes are 22000 and 35000 respectively. The tokenizer of \textsc{GPT2} is shared with \textsc{LSTM}.

\paragraph{Syntax Extracting Methods} To evaluate the branching bias, we use the codes\footnote{\url{https://github.com/galsang/trees_from_transformers} and \url{https://github.com/LividWo/Perturbed-Masking}} of \citet{kim2020pre} and \citet{wu2020perturbed}, and re-implement the algorithm in \citet{marevcek2018extracting}. The parsing algorithms proposed by them are referred to as \textsc{Dist}, \textsc{Mart}, and \textsc{AttnSpan} respectively. Note that \citet{kim2020pre} propose a trick to explicitly inject right-branching bias to their method, and we set the weight of this injected external bias to zero in our experiments. For feature definitions, we mainly focus on  three types of feature definitions, which are hidden representation~\cite{kim2020pre}, full attention ~\cite{marevcek2018extracting}, and prefix attention ~\cite{kim2020pre,wu2020perturbed}.~\footnote{Prefix-attention means the attention is performed over the prefix words as in \textsc{GPT2} whereas full-attention is over all words in a sentence as in \textsc{BERT}.} The hyper-parameters (e.g., choices of attention head and hidden layer) of syntax extracting methods are tuned on the validation set.

\subsection{Main Results}

As shown in Table \ref{tab:pipeline}, the behaviors of different approaches are widely divergent. We find that the branching bias in \textsc{BERT}+\textsc{AttnSpan} and \textsc{BERT}+\textsc{Dist} are relatively lower than other approaches. However, the results of \textsc{GPT2}+\textsc{AttnSpan} and \textsc{BERT/GPT2}+\textsc{Mart} demonstrate significant right-branching biases. \textsc{GPT2}+\textsc{Dist} shows a tendency towards left-branching. Since these approaches are pipelined, which part of their methods has an impact on the branching bias is still unclear.

The results reported in Table \ref{tab:pipeline} is a little worse than those reported in \citet{kim2020pre, wu2020perturbed}. One reason is that we evaluate the results on the corpus-level F1 score following the standard, rather than sentence-level \cite{kim2020pre}. The other reason is that our training data is small, since it is too expensive to train reversed language models on a huge dataset. However, these results are obtained by running the released codes of \citet{kim2020pre,wu2020perturbed},
and thus, we think it will not affect our findings.


\subsection{Factors affecting branching bias\label{sec:3.3}}

\paragraph{Branching Bias in Parsing Algorithm} The branching gaps of different parsing algorithms are shown in Table \ref{tab:parser}. Observing from the experiment results in English, The branching gaps of \textsc{Mart} is significantly larger than 0, which means it has a tendency to right-branching. In contrast, the branching gaps of parsing algorithm \textsc{AttnSpan} and \textsc{Dist} are nearly 0, which means they do not bias to left-branching or right-branching. Although \textsc{Dist} is inspired by the parsing algorithm in \citet{shen2018ordered}, it is an unbiased, which is consistent with the claim in \citet{kim2020pre}. We also evaluate the parsing algorithm of \citet{shen2018ordered}, and its branching gap is $+3.22$ on English, which is consistent with the finding in \citet{dyer2019critical}.

To examine whether some other language properties might play a role in this process, we also conduct experiments on different languages, which can help rule out the impact of specific language properties. The results in Table \ref{tab:parser} show that \textsc{Mart} has the same trend as \textsc{Right-B} baseline (row 5) on both Chinese and German datasets, which is consistent with the finding on the English dataset. It is also worth noting that the branching gap for \textsc{Mart} is positive on Chinese and English datasets, whereas it is negative on German. The reason is that both Chinese and English are right-branching languages, while German is inclined to be left-branching. However, both head-initial and head-final structures occur in the German language from the viewpoint of linguistics.  In addition, one interesting observation is that, the performances of \textsc{AttnSpan} are always higher than the \textsc{Random} baseline. We hypothesize that \textsc{AttnSpan} may have a bias towards the balance tree, due to its way to compute weights of splitting points.



\paragraph{Branching Bias in Feature Definition} As shown in Table \ref{tab:definition}, we choose the unbiased parsing algorithm \textsc{Dist} to further analyze the branching bias in feature definitions. It is worth noting that, after normalization, the attention matrix of \textsc{Prefix-Attn} is lower triangular, and that of \textsc{Full-Attn} is fully filled. We find that the feature definitions based on \textsc{Hidden} and \textsc{Full-Attn} are unbiased. However, \textsc{Prefix-Attn} tends to generate right-branching trees, where the branching gap is $+7.27$ on English. This finding is consistent with that on Chinese and German. One possible explanation about \textsc{Prefix-Attn} is that the attention scores will become distracted with the prefix grows, such that the feature scores in the front of the sequence, which has a larger value, would be picked at first.

\begin{table}[t]
  \begin{center}
  \resizebox{1\columnwidth}{!}{
  \begin{tabular}{c|c||c|c|c}\
\#  & Language Model & $L$ & $L^\prime$ & Gap \\\hline
1 & \textsc{BERT}   &   $24.93$   &     $25.23$   & $-0.30$   \\
2 &   \textsc{GPT2}   &   $23.85$    &     $26.09$  & $-2.24$   \\
3 &   \textsc{LSTM} &   $28.72$   &   $26.22$  & $+2.50$ \\\hline
4 &  \textsc{Random}   & $18.31$  &    $18.37$  & $-0.06$ \\
  \end{tabular}
  }
  \end{center}
  \caption{\label{tab:model} The branching gaps of different language models using an unbiased pipeline \textsc{Hidden} + \textsc{Dist}. The results are corpus-level F1 scores on English.}
  \end{table}

\paragraph{Branching Bias in Language Models}
After the analyses in previous steps, we will use the unbiased parsing algorithm and feature definition, \textsc{Dist} and \textsc{Hidden}, to evaluate the branching bias in language models. Note that the results in this section is different from those in Table \ref{tab:pipeline}, since other feature definitions are prohibited except for \textsc{Hidden}.

Our experiments conducted on language models are shown in Table \ref{tab:model}. The performances of \textsc{BERT} on both branching are nearly the same, where the branching gap is just $-0.30$. In contrast, slight branching gap is observed on both \textsc{GPT2} and \textsc{LSTM}. The branching gap of \textsc{GPT2} is $-2.24$. With the same left-to-right paradigm, \textsc{LSTM} behaviors a positive branching gap $+2.50$. The opposite branching gap may be caused by the difference between model architectures, where \textsc{GPT2} is based on self-attention \cite{vaswani2017attention} and \textsc{LSTM} is based on gating mechanism \cite{hochreiter1997long}. However, the random noises may also play a role in this observation, since the performance range of \textsc{GPT2} models trained on the original English dataset with different random seed can also reach around $1.50$. We will investigate it in future works.


\section{Conclusion}

In this paper, we propose an approach to quantitatively analyze the branching bias in extracting syntax from pre-trained language models. Unlike previous work, our approach is more general to be applied to any pre-trained language models and syntax extracting methods. Furthermore, we systematically analyze three factors in depth that may affect the branching bias: the language model, feature definition, and parsing algorithm. Our experiments show that branching biases are in many recent works. In addition, these biases can be brought by each of the three factors. We appeal that researchers should carefully design their syntax extracting method to reveal the real competence of syntax from a pre-trained language model.

\bibliographystyle{acl_natbib}
\bibliography{emnlp2020}
\end{document}